\documentclass[10pt,twocolumn,letterpaper]{article}

\usepackage{cvpr}
\usepackage{times}
\usepackage{epsfig}
\usepackage{graphicx}
\usepackage{amsmath}
\usepackage{amssymb}
\usepackage{multirow}
\usepackage{array}
\usepackage{subfigure}

\usepackage[pagebackref=true,breaklinks=true,letterpaper=true,colorlinks,bookmarks=false]{hyperref}

\cvprfinalcopy 

\def\cvprPaperID{1618} 

\ifcvprfinal\fi
\begin{document}

\title{Geometry Guided Adversarial Facial Expression Synthesis}

\author{Lingxiao Song$^{1,2}$ \quad Zhihe Lu$^{1,3}$ \quad Ran He$^{1,2,3}$ \quad Zhenan Sun$^{1,2}$ \quad Tieniu Tan$^{1,2,3}$\\
$^{1}$National Laboratory of Pattern Recognition, CASIA\\
$^{2}$Center for Research on Intelligent Perception and Computing, CASIA\\
$^{3}$Center for Excellence in Brain Science and Intelligence Technology, CAS\\
}

\maketitle

\begin{abstract}
Facial expression synthesis has drawn much attention in the field of computer graphics and pattern recognition. It has been widely used in face animation and recognition. However, it is still challenging due to the high-level semantic presence of large and non-linear face geometry variations. This paper proposes a Geometry-Guided Generative Adversarial Network (G2-GAN) for photo-realistic and identity-preserving facial expression synthesis. We employ facial geometry (fiducial points) as a controllable condition to guide facial texture synthesis with specific expression. A pair of generative adversarial subnetworks are jointly trained towards opposite tasks: expression removal and expression synthesis. The paired networks form a mapping cycle between neutral expression and arbitrary expressions, which also facilitate other applications such as face transfer and expression invariant face recognition. Experimental results show that our method can generate compelling perceptual results on various facial expression synthesis databases. An expression invariant face recognition experiment is also performed to further show the advantages of our proposed method.

\end{abstract}

\section{Introduction}

Facial expression synthesis is a classical graphics problem where the goal is to generate face images with specific expression for specified human subject. It has drawn much attention in the field of computer graphics, computer vision and pattern recognition. Synthesizing photo-realistic facial expression images has been of great value for both academic and industrial communities, and has been widely applied in facial animations, face editing, face data augmentation and face recognition. During the last two decades, many facial expression synthesis methods have been proposed, which can be roughly divided into two categories. The first category mainly resorts to computer graphics technique to directly warp input faces to target expressions~\cite{zhang2006geometry,yang2012facial,Yeh2016Semantic} or re-use sample patches of existing images~\cite{Mohammed2009Visio}, while the other aims to build generative models to synthesize images with predefined attributes~\cite{susskind2008generating,Ding2017ExprGAN}.

For the first category, a lot of research efforts have been devoted to finding correspondence between existing facial textures and target images. Earlier approaches usually generate new expressions by creating fully textured 3D facial models~\cite{pighin2006synthesizing,Blanz2003Reanimating}, warping face images via feature correspondence~\cite{theobald2009mapping} and optical flow~\cite{yang2012facial,yang2011expression}, or compositing face patches from an existing expression dataset~\cite{Mohammed2009Visio,Kemelmacher2010Being}. Particularly, Yeh et al.~\cite{Yeh2016Semantic} propose to learn the optical flow with a variational autoencoder. Although this kind of methods can usually produce realistic images with high resolution, their elaborated yet complex processes often result in expensive computation.

The representative methods in the second category are deep generative models that have recently obtained impressive results for image synthesis applications~\cite{zhu2017unpaired,pix2pix2016,huang2017beyond}. However, images generated by such methods sometimes lack fine details and tend to be blurry or of low-resolution. Targeted expressional attributes are usually encoded in a latent feature space, where certain directions are aligned with semantic properties. Therefore, these methods can provide better flexibility in semantical-level image generation, but it is hard to take fine-grain control of the synthesized images, e.g., widen the smile or narrow the eyes.

\begin{figure*}[t]
\begin{center}
\includegraphics[width=0.9\linewidth]{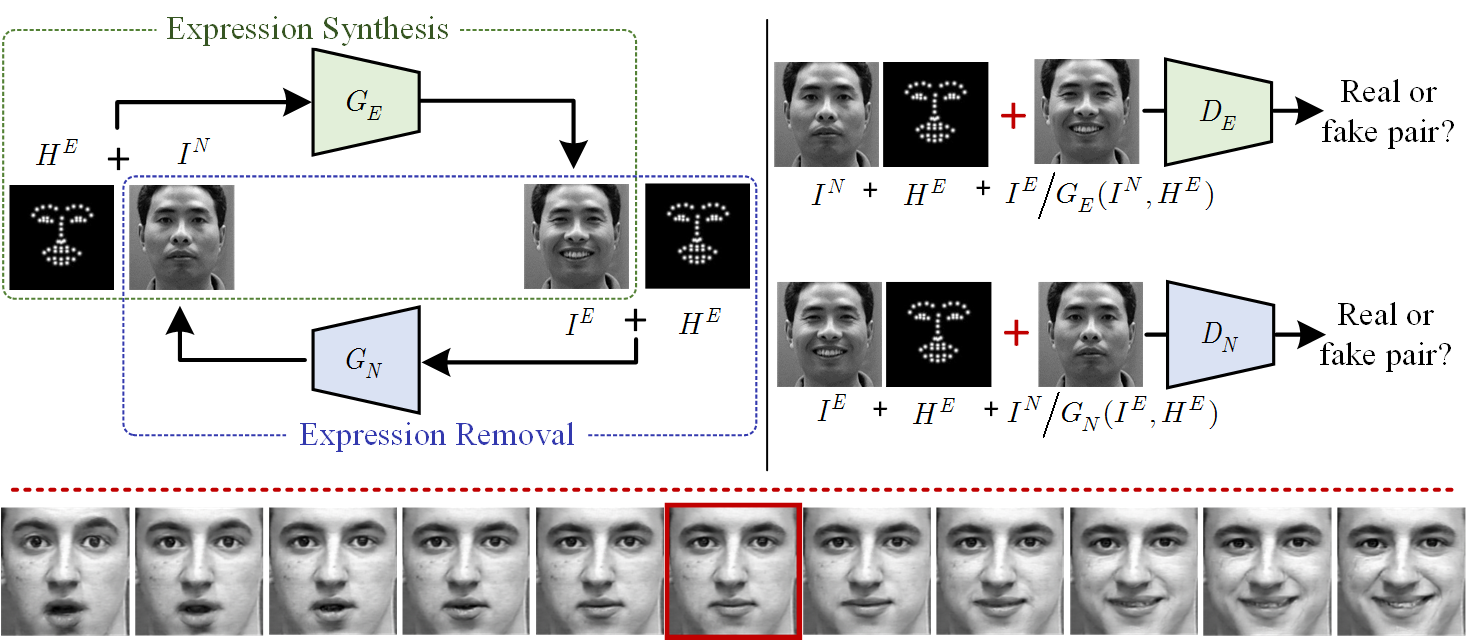}
\end{center}
   \caption{The proposed geometry-guided facial expression generation framework. Face geometry is fed into generators as the condition to guide the processes of expression synthesis and expression removal. In the bottom, we show some examples generated from the same real face image (the center one marked with red box).}
\label{fig:pipeline}
\end{figure*}

In this paper, a deep architecture (G2-GAN) is proposed to synthesize photo-realistic and identity-preserving facial images while keep operation-friendly. A human face is often assumed to contain geometry and texture information~\cite{Matthews2004Active} in computer vision, and both geometry and texture  attributes can be used to facilitate face recognition and expression classification~\cite{li2006expression}. Inspired by the face geometry information in active appearance models (AAM), we employ face geometry to control the expression synthesis process.  Face geometry is defined via a set of feature points, and is transformed to an image (heat map) and fed to G2-GAN as a control condition. Fig.~\ref{fig:pipeline} is the pipeline of our approach. We generate facial expression images conditioned on both the input face images and geometry attributes. Particularly, expression generating and removal are simultaneously considered in our method, constructing a full mapping cycle of expression editing. Then, expression transfer can be performed between arbitrary expressions and subjects. Extensive experiments on two facial expression databases demonstrate the superiority of the proposed facial expression synthesis framework.

The main contributions are summarized as follows,
\begin{itemize}
\item We propose a novel geometry-guided GAN architecture for facial expression synthesis. It can generate photo-realistic images in different expressions from a single image, where target expressions can be easily controlled by various facial geometry inputs.
\item We employ a pair of GANs to simultaneously perform two opposite tasks: removing expression and synthesizing expression. By combining these two models, our method can be used in many applications such as facial expression transfer and cross-expression recognition.
\item We utilize an individual-specific shape model to operate facial geometry, which gives consideration to individual differences when perform expression synthesis. Based on this model, facial expression transfer and interpolation can be easily conducted.
\item Extensive experiments on two facial expression databases demonstrate that the proposed method can synthesize photo-realistic and identity-preserving expression images.
\end{itemize}

\section{Related Works}

Facial expression synthesis (or editing) is an important task in face editing. In this section, we briefly review some recent advances in facial expression synthesis and its related generative adversarial networks (GAN).

\subsection{Expression synthesis}

As mentioned above, existing expression synthesis methods can be categorized to two classes according to the way of manipulating pixels.

Methods in the first category address this problem either with 2D/3D image warping~\cite{Blanz2003Reanimating,garrido2014automatic}, flow mapping~\cite{yang2011expression,Yeh2016Semantic} or image reordering~\cite{li2012data,yang2012facial}, most of which are morph-based or example-based.
For instance, ~\cite{Blanz2003Reanimating} estimates 3D shape from a neutral face, and synthesizes facial expression by 3D rendering.
Bolkart et al.~\cite{bolkart2015groupwise} propose a groupwise multilinear correspondence optimization to iteratively refine the correspondence between different 3D faces.
In~\cite{garrido2014automatic}, an image-based warping strategy is introduced to perform automatic face reenactment, with the facial identity preserving being considered.
Thies et al.~\cite{Thies2016Face2Face} track expressions based on a statistical facial prior, and then achieve real-time facial reenactment by using deformation transfer in a low-dimensional expression space.
Particularly, Olszewski et al.~\cite{olszewski2017realistic} 
employ a generative adversarial framework to refine 3D texture correspondences and infer details such as wrinkles and inner mouth region.
Many works attempt to utilize the optical flow map to perform image warping.
In~\cite{yang2011expression}, 3D faces of different expressions are constructed, and expression flow is computed by projecting the difference between 3D shapes back to 2D.
Recently, neural networks based methods~\cite{ganin2016deepwarp,Yeh2016Semantic} have been presented to manipulate expression flow maps. It is difficult for those warping-based methods to recover unseen facial components, e.g., skin wrinkles and inner mouth area, or synthesize realistic images for new faces.

Example-based methods edit faces by re-using image patches or reordering image samples of a training set, which can synthesize expected expressions as well as generate unseen faces. \cite{Mohammed2009Visio} composites facial patches from a large dataset to synthesize face images with desired expressions.
In~\cite{Kemelmacher2010Being}, expression is mapped to a new face by matching images with similar pose and expression from a database of the target person.
Li et al.~\cite{li2012data} hallucinate face videos by retrieving frames via a carefully-designed expression similarity metric from an existing expression database. Yang et al.~\cite{yang2012facial} reorder input face frames using Dynamic Time Warping, and then apply an additional expression warping to get more realistic results.

The other kinds of methods use generative models to deal with facial expression synthesis. In~\cite{susskind2008generating}, a deep belief net is used to convert high-level descriptions of facial attributes into realistic face images.
Reed et al.~\cite{reed2014learning} propose a higher-order Boltzmann machine to model interaction among multiple groups of hidden units, and each unit group encodes distinct variation factors such as pose, morphology and expression in face images.
In~\cite{cheung2014discovering}, a regularization term is embedded in an autoencoder to disentangle the variations between identity and expression in an unsupervised manner.
Li et al.~\cite{li2016deep} build a convolutional neural network to generate facial images with the given source input images and reference attribute labels.
Shu et al.~\cite{NeuralFace2017} learn a face-specific disentangled representation of intrinsic face properties via GAN, and generate new faces by changing the latent representations.
Recently, Ding et al.~\cite{Ding2017ExprGAN} propose ExprGAN which can synthesize facial expression with controllable intensity, and an expression controller network is proposed to learn expression code. ExprGAN is the most similar work to ours as far as we know. However, ExprGAN generates images conditioned on expression labels and intensity values, while we employ the face geometry as control condition which is not limited to certain expression styles.    

\subsection{Generative adversarial networks}
Our work is also related to the generative adversarial networks (GAN)~\cite{goodfellow2014generative}, which provides a simple yet efficient way to train powerful model via the min-max two player game between generator and discriminator. Many modified architectures of GAN have been proposed to deal with different tasks. For example, CGAN\cite{mirza2014conditional} introduces a conditional version of GAN to guide image synthesis process via adding supervised information to both generator and discriminator. CycleGAN~\cite{zhu2017unpaired}, DualGAN~\cite{yi2017dualgan} and DiscoGAN~\cite{kim2017learning} share the same idea of employing a cycle structure to handle the unpaired image-to-image translation problem.
GAN and its variants have achieved great success in numerous image-generating-related tasks such as image synthesis~\cite{radford2015unsupervised}, image super-resolution~\cite{ledig2017photo}, image style transfer~\cite{zhu2017unpaired,pix2pix2016} and face synthesis~\cite{huang2017beyond}. Motivated by this, we develop our facial expression synthesis framework based on GAN, aiming at generating photo-realistic images with high-quality local details.

\section{Methods}
In this section, we present a novel framework for the facial expression synthesis problem based on generative adversarial networks.
We first describe the geometry guided facial expression synthesis in detail, and then propose geometry manipulation methods for face transfer and expression interpolation.


\subsection{Geometry Guided Facial Expression Synthesis}
The outstanding performance of GAN in fitting data distribution has significantly promoted many computer vision applications such as image style transfer~\cite{zhu2017unpaired,pix2pix2016}. Motivated by its remarkable success, we employ GAN to perform the facial expression synthesis.

Only limited expression styles are supported by existing deep learning-based facial expression synthesis methods, which are usually semantic properties such as smile and angry.
Many works can transform a neutral face to a smile face, but can hardly control how strong the smile is. Even though one can construct an intensity-sensitive model by using training data with emotion intensity annotations, many expressions are still difficult to encode with the limited semantic properties. For example, it is hard to describe ``a lopsided grin with one eye open" using normal semantic properties. To address this problem, we employ the face geometry to guide the generation.

As in AAM, face geometry is defined via a set of fiducial points~\cite{Matthews2004Active}.
Heatmap is used to encode the locations of these facial fiducial points, which has been widely used in human pose estimation~\cite{tompson2014joint} and face alignment~\cite{huang2015densebox}. The heatmap provides a per-pixel likelihood for fiducial point locations. Given the heatmaps of target facial expressions and frontal-looking faces without expression (in the following we term it as expressionless faces), new face images (expressioned faces) are synthesized accordingly.

As illustrated in Fig.~\ref{fig:pipeline}, a pair of generators ${G_E}:{(I^N,H^E)} \to {I^E}$ and ${G_N}:{(I^E,H^E)} \to {I^N}$ are introduced, in which $I^N$ is an expressionless face, $I^E$ is an expressioned face and $H^E$ is the heatmap corresponding to $I^E$. Associated with these two generators, two discriminators $D_E$ and $D_N$ are involved, aiming to distinguish between real triplets $(I,H,I')$ and generated triplets $(I,H,G(I))$ correspondingly. $I$ and $I^{'}$ are images of expressionless and expressioned faces, or vise versa.

It is worth noting that $H^E$ plays different roles in these two face editing models, i.e., control measure in expression synthesis and auxiliary annotation in expression removal.
In the expression synthesis process, $H^E$ is used to specify the target expression so that $G^E$ can transform neutral expression $I^N$ into desired expression.
As for the expression removal process, $H^E$ is in charge of indicating the state of $I^E$ so as to facilitate the recovering of $I^N$.

\textbf{Adversarial Loss}.
Generators and discriminators are trained alternatively towards adversarial goals, following the pioneering work of~\cite{goodfellow2014generative}. Since the proposed face editing models generate results conditioned on the input face images and heatmaps, we apply GAN in conditional setting as~\cite{mirza2014conditional,pix2pix2016}. The adversarial losses for generator and discriminator are shown in Eq.~\ref{L_adv_G} and Eq.~\ref{L_adv_D} respectively.
\begin{equation}\label{L_adv_G}
\begin{split}
{L_{G - adv}} =  - {{\mathbb{E}}_{I,H \sim P\left( {I,H} \right)}}\log D\left( {I,H,G\left( {I,H} \right)} \right)
\end{split}
\end{equation}
\begin{equation}\label{L_adv_D}
\begin{split}
{L_{D - adv}} &= {{\mathbb{E}}_{I,H,I' \sim P\left( {I,H,I'} \right)}}\log \left( {1 - D\left( {I,H,I'} \right)} \right)\\
&+ {{\mathbb{E}}_{I,H \sim P\left( {I,H} \right)}}\log D\left( {I,H,G\left( {I,H} \right)} \right)
\end{split}
\end{equation}

\textbf{Pixel Loss}.
The generator is tasked to not only fool the discriminator, but also synthesize images similar to the target ground-truths as far as possible. The pixel-wise loss $L_{pixel}$ enforces the transformed face image to have a small distance with the ground-truth in the raw-pixel space. $L_{pixel}$ takes the form:
\begin{equation}\label{L_pixel}
\begin{split}
{L_{pixel}} = {E_{I,H,I' \sim P\left( {I,H,I'} \right)}}{\left\| {I' -  {G\left( {I,H} \right)}} \right\|_1},
\end{split}
\end{equation}
where we use L1 distance to encourage less blurring output. $(I,H,I')$ is one of the combination of $(I^N,H^E,I^E)$ and $(I^E,H^E,I^N)$ depending on the generators.

\textbf{Cycle-Consistency Loss}.
The generators $G_E$ and $G_N$ construct a full mapping cycle between neutral expression faces and expressioned faces. If we transform a face image from neutral expression to angry and then transform it back to neutral expression, the same face image should be obtained in the ideal situation. Therefore, we introduce an extra cycle consistency loss $L_{cyc}$ to guarantee the consistency between source images and the reconstructed images, e.g., $I^N$ vs. ${G_N}\left( {{G_E}\left( {{I^N},{H^E}} \right),{H^E}} \right)$ and $I^E$ vs. ${G_E}\left( {{G_N}\left( {{I^E},{H^E}} \right),{H^E}} \right)$. $L_{cyc}$ is calculated as
\begin{equation}\label{L_cyc}
\begin{split}
{L_{cyc}} = {E_{I,H \sim P\left( {I,H} \right)}}{\left\| {I - G'\left( {G\left( {I,H} \right)} \right)} \right\|_1},
\end{split}
\end{equation}
where $G'$ is the opposite generator to $G$. In our case, if $G$ is used to transform neutral expression into expression specified by the face geometry heatmap $H$, then $G'$ is used to recover the neutral expression with the assistance of $H$.

\textbf{Identity Preserving Loss}.
A fundamental principle of facial expression editing is that face identity should be preserved after expression synthesis as well as removal. Thus, an identity-preserving term is adopted in our framework to enforce identity consistency:
\begin{equation}\label{L_identity}
\begin{split}
{L_{identity}} = {E_{I,H \sim P\left( {I,H} \right)}}{\left\| {F\left( I \right) - F\left( {G\left( {I,H} \right)} \right)} \right\|_1},
\end{split}
\end{equation}
where $F$ is a feature extractor for face recognition.
We employ the model-B of the Light CNN~\cite{wu2015lightened} as our feature extraction network, which includes 9 convolution layers, 4 max-pooling layers and one fully-connected layer. The Light CNN is pre-trained as a classier to distinguish between tens of thousands of identities, so it has ability to capture the most prominent feature for face identity discrimination. Therefore, we can leverage this loss to enforce preserving face identity through the face editing processes.


To sum up, the final full objective for generators $G_N, G_E$ is a weighted sum of all the losses defined above: $L_{G - adv}$ to remove the modality gap between real and generated samples, $L_{pixel}$ to force pixel-wise correctness, $L_{cyc}$ to guarantee cycle consistency of the reconstructed image and source image, and $L_{identity}$ to preserve identity characteristic through mapping process.
\begin{equation}\label{L_G_full}
\begin{split}
{L_G} = {L_{G - adv}} + {\alpha _1}{L_{pixel}} + {\alpha _2}{L_{cyc}} + {\alpha _3}{L_{identity}}
\end{split}
\end{equation}
where $\alpha_1$,$\alpha_2$ and $\alpha_3$ are loss weight coefficients.

\subsection{Facial Geometry Manipulation}
\label{sec:Geometry_Manipulation}
As mentioned above, geometric positions of a set of fiducial points are employed to guide facial expression editing in our framework. Face geometry is largely affected by facial expression, and is a useful cue for expression recognition~\cite{li2006expression}. Its usage provides a more intuitive yet efficient way for specifying target facial expression. This is because face geometry can not only visually represent the locations and shapes of facial organs, but also be adjusted continuously to obtain expressions with different intensities.

Human faces have unique physiological structure characteristics, resulting in strong correlation between the locations of fiducial points. Hence, the variance of facial geometry should be constrained to avoid unreasonable settings, e.g., eyebrows under the eyes, square-shapes eyes or nose.
Taking the prior knowledge of faces' distribution into account, a parametric shape model is built to serve as a geometry generator.

We adopt a method similar to~\cite{Matthews2004Active} to learn a basic shape model from labelled training images. Firstly, faces are normalized to the same scale and rotated to horizontal according to the locations of two eyes. Then, Principal Component Analysis (PCA) is applied 
to get a basic shape model of the locations for $K$ fiducial points
\begin{equation}\label{Eq:Shape}
\begin{split}
s(p) = {s_0} + Sp
\end{split}
\end{equation}
where $s, {s_0} \in {R^{2K \times 1}}$ , $S \in {R^{2K \times N}}$, $p \in {R^{N \times 1}}$. The base shape $s_0$ is the mean shape of all the training images and columns of $S$ are the $N$ eigenvectors corresponding to the $N$ largest eigenvalues. Different facial geometries can be obtained by changing the value of shape parameters $p$. 

However, facial geometry is not only correlated with facial expression, but also related to face identity to a great extent. The facial geometry varies with different individuals even under the same expression. For example, the distance between eyes and the length of nose depend largely on face identity rather than expression. Considering these individual differences, we propose an individual-specific shape model based on Eq.~\ref{Eq:Shape}, which can be
derived by replacing the mean shape $s_0$ with the neutral shape $s_0^I$ of different individuals.
The individual-specific shape model is given by
\begin{equation}\label{Eq:Shape_individual}
\begin{split}
s^I(p) = {s_0^I} + Sp
\end{split}
\end{equation}
where $s_0^I$ accounts for variation relate to identity, while $p$  accounts for changes caused by facial expression.

\textbf{Facial Expression Transfer}.
The proposed framework can be easily applied in facial expression transfer. Given two expressioned faces $I^A$ and $I^B$ with detected facial landmarks $s^A, s^B$. The expression removal model is firstly employed to recover expressionless faces as
\begin{equation}\label{Eq:expression_removing1}
\begin{split}
I_0^A = G_N(I^A,s^A),\ I_0^B = G_N(I^B,s^B)
\end{split}
\end{equation}
where $I_0^A,I_0^B$ denote the neutral expression faces of $I^A, I^B$ respectively. Therefore the neutral shapes $s_0^A, s_0^B$ can be acquired via facial landmark detection.

Then, the shape parameters are derived by solving the following least squares regression problem.
\begin{equation}\label{Eq:Shape_transfer1}
\begin{split}
{p^A} = \arg \mathop {\min }\limits_p {\left\| {{s^A} - s_0^A - Sp} \right\|^2}\\
{p^B} = \arg \mathop {\min }\limits_p {\left\| {{s^B} - s_0^B - Sp} \right\|^2}
\end{split}
\end{equation}

We change shape parameters so as to get transferred locations of fiducial points.
\begin{equation}\label{Eq:Shape_transfer2}
\begin{split}
{s^{AB}} = s_0^A + S{p^B}\\
{s^{BA}} = s_0^B + S{p^A}
\end{split}
\end{equation}
Heatmaps are transformed according to these transferred shapes, and concatenated with corresponding expressionless faces as inputs for expression synthesis.
Finally, results of facial expression transfer can be obtained by using our expression synthesis model as Eq.~\ref{Eq:expression_removing3}.
\begin{equation}\label{Eq:expression_removing3}
\begin{split}
I^{AB} = G_E(I_0^A,s^{AB}),\ I^{BA} = G_E(I_0^B,s^{BA})
\end{split}
\end{equation}

\textbf{Facial Expression Synthesis and Interpolation}.
As mentioned above, our method is able to synthesize different expressions from a single image. The simple requirement is to prepare a neutral expression face image and shape parameters for target expression. Benefitting from the proposed expression removal model, neutral expression face is not hard to access. The shape parameters for specific expression can be learnt via the basic shape model (see in Eq.~\ref{Eq:Shape}) from annotated training dataset. Once the values of shape parameters are associated with certain semantic properties, such as fear and surprise, we can use them to synthesize unseen facial expressions with desired semantic types. Besides, facial expression interpolation can be conducted by linearly adjusting the value of shape parameters.


\section{Experiments}
In this section, we evaluate the proposed approach on two commonly used facial expression databases. The databases and testing protocols are introduced firstly.
Then, the implementation details are presented. Finally, we provide experiments with qualitative and quantitative results for single-image editing, face transfer, expression interpolation and expression-invariant face recognition.

\subsection{Datasets and Protocols}
\textbf{The CK+ database}~\cite{lucey2010extended}.
CK+ database includes 593 sequences from 123 subjects, in which seven kinds of emotions are labeled. The first frame is always neutral while the last frame has the peak expression. In each expression video sequence, the first frame is selected as the neutral expression, while last half frames are used as target expression. Training and testing subsets are divided based on identity, with 100 for training and 23 for testing. Locations of 68 fiducial points of each frame are provided, and we use them to create heatmaps for experiments. Because almost all of the videos in CK+ database are grayscale, grayscale images are used in our experiment.

\textbf{The Oulu-CASIA NIR-VIS facial expression database~\cite{chen2009learning}}.
Videos of 80 subjects with six typical expressions and three different illumination conditions are captured in both NIR and VIS imaging systems in this database. Only images captured by a VIS camera within strong illumination condition are used in our facial expression editing experiments.
Similar to the CK+ database, we take the first frame and images belong the last half of each sequence to make training pairs.
The Oulu-CASIA database includes two parts captured among different ethnic groups at different time, where P001 to P050 are Finnish people and the rest P051 to P080 are Chinese people. We find that these two parts differ a lot in illuminations and face structures. Hence, we select training data over these two parts. Finally we get a training subset of 60 subjects that consists of 37 Finns and 13 Chinese, and a testing subset with 13 Finns and 7 Chinese accordingly. We use the 68 fiducial points detected by~\cite{bulat2017far} to create heatmaps.


\subsection{Implementation Details}
\textbf{Image pre-processing}.
All the face images are normalized by the similarity transformation using the locations of two eyes, and then cropped to $144 \times 144$ size, of which $128 \times 128$ sized sub images are selected by random cropping in training and center cropping in testing. In training stage, we also perform random flipping of the input images to encourage generalization performance.
The heatmap is a multi-channel image with the same size as input face image, where value of each pixel is the likelihood for fiducial point location. 2D Gaussian convolution is applied on each channel to smooth the heatmap.
All the pixel values are normalized into range of [0,1], including face images and heatmaps.


\textbf{Network architecture}.
We adapt our architecture from~\cite{pix2pix2016}. The generators take the architecture of U-Net, which is an encoder-decoder with skip connections between mirrored layers in the encoder and decoder stacks.
For discriminator networks, the frequently-used PatchGAN model is employed.

We train different models for each dataset with a batch size of 5 and an onset learning rate of $10^{-4}$.
In all our experiments, hyper-parameters are set empirically to balance the importance of different losses. The trade-off parameter $\alpha _1$ for pixel loss is set to 10, $\alpha_2$ for cycle consistency loss is set to 5. $\alpha_3$ for identity-preserving loss is set to 0.1 in the beginning, and is gradually increased to 0.5 along with the training process.

\subsection{Experimental Results}

\subsubsection{Facial Expression Editing}

\begin{figure}[t]
\begin{center}
\includegraphics[width=1.0\linewidth]{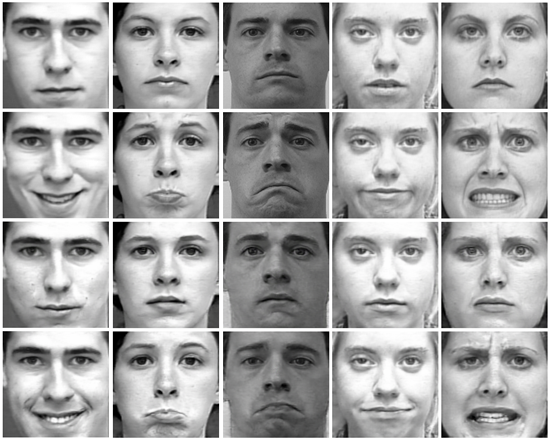}
\end{center}
   \caption{Results of CK+ database for facial expression synthesis and removal. From top to bottom, input expressionless images (true $I^N$), input expressioned images (true $I^E$), expression removal results (fake $I^N$) and expression synthesis results (fake $I^E$).}
\label{fig:res_syn}
\end{figure}

\begin{figure}[t]
\begin{center}
\includegraphics[width=1.0\linewidth]{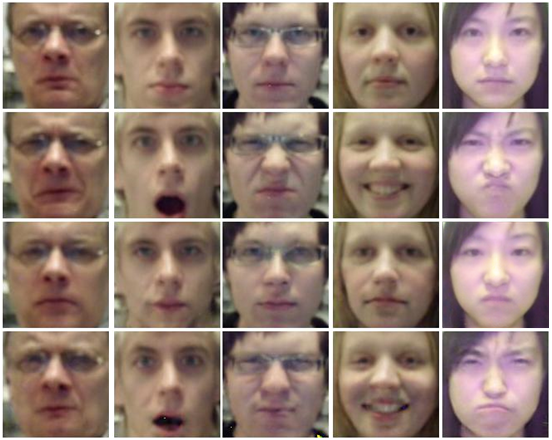}
\end{center}
   \caption{Results of Oulu-CASIA database for facial expression synthesis and removal. Images are arranged by the same order as Fig.~\ref{fig:res_syn}.}
\label{fig:res_syn_oulu}
\end{figure}

For this experiment, given testing image triplets $(I^N, I^E, H^E)$, we conduct expression synthesis on $(I^N,H^E)$ and expression removal on $(I^E,H^E)$ simultaneously.
Some visual examples are shown in Fig.~\ref{fig:res_syn} and Fig.~\ref{fig:res_syn_oulu}. The first two rows display original expressionless faces and original expressioned faces, and the next two rows are results of expression removal and expression synthesis respectively. We can see that the proposed G2-GAN is capable of generating compelling identity-preserving faces for desired expression in both testing datasets.
Since the images in the CK+ database have higher resolution than those in the Oulu-CASIA database, results for the CK+ database contain better low-level image quality such as skin wrinkles.
Noting that we can synthesize satisfactory mouth region with even teeth textures, without needing to involve extra manipulations such as recovering mouth area by retrieving similar frames from a pre-trained database.

In order to measure the correctness of transformed images, we adopt PSNR (peak signal to noise ratio, dB) and SSIM (structural similarity index) for quantitative metric, where PSNR is calculated on the luminance channel and SSIM is calculated on three channels of RGB respectively. Tab.~\ref{tab:res_psnr} reports quantitative results of the proposed approach under different settings. Both the cycle consistency loss and the identity preserving loss contribute to improve performances, and the best result is acquired by combining them together.



\begin{table*}[htbp]
  \centering
  \caption{Quantitative results for expression synthesis and expression removal on CK+ and Oulu-CASIA databases.}
    \begin{tabular}{|p{1.9cm}<{\centering} | p{4.0cm}<{\centering} | p{1.8cm}<{\centering} | p{1.8cm}<{\centering} |  p{1.8cm}<{\centering} | p{1.8cm}<{\centering}|}
    \hline
     \multirow{2}[2]{*}{Dataset}&\multirow{2}[2]{*}{Configuration} & \multicolumn{2}{c|}{Expression Removal} & \multicolumn{2}{c|}{Expression Synthesis} \\
     \cline{3-6}
           & & SSIM  & PSNR  & SSIM  & PSNR \\
    \hline
    \multirow{4}[2]{*}{CK+} & w/o $L_{cyc}$, $L_{identity}$ & 0.726 & 22.655 & 0.756 & 23.903 \\
          & w/o $L_{cyc}$ & \textbf{0.728} & 22.828 & 0.754 & 24.061 \\
          & w/o $L_{identity}$ & 0.724 & 22.516 & 0.765 & 24.335 \\
          & G2-GAN & \textbf{0.728} & \textbf{22.968} & \textbf{0.767} & \textbf{24.420} \\
    \hline
    \multirow{4}[2]{*}{Oulu-CASIA} & w/o $L_{cyc}$, $L_{identity}$ & 0.902 & 25.202 & 0.908 & 26.206 \\
          & w/o $L_{cyc}$ & 0.903 & 25.270 & 0.914 & 26.337 \\
          & w/o $L_{identity}$ & 0.904 & 25.519 & \textbf{0.916} & \textbf{26.677} \\
          & G2-GAN & \textbf{0.910} & \textbf{25.810} & 0.914 & 26.588 \\
    \hline
    \end{tabular}%
  \label{tab:res_psnr}%
\end{table*}%


\subsubsection{Facial Expression Transfer}
In this part, we demonstrate our model's ability to transfer the expression of different faces. The procedures for facial expression transfer are introduced in Sec.~\ref{sec:Geometry_Manipulation}.

\begin{figure}[t]
\begin{center}
\includegraphics[width=0.95\linewidth]{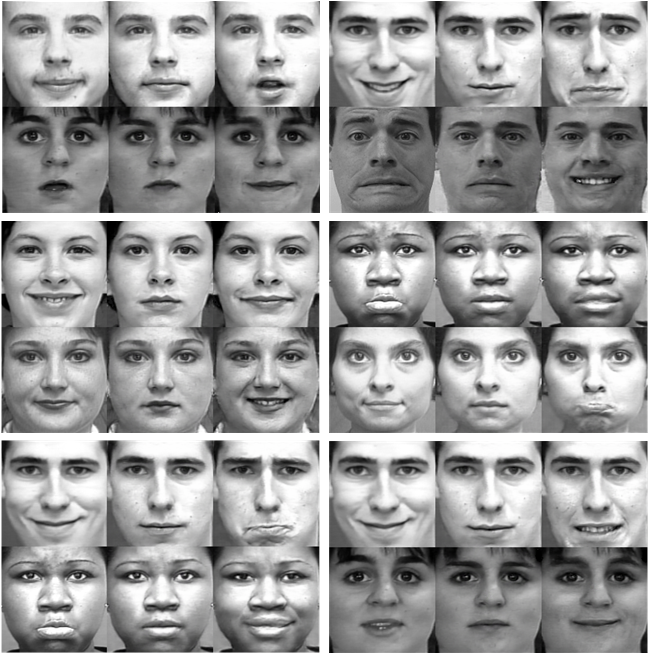}
\end{center}
   \caption{Results of CK+ database for facial expression transfer. There are three images for each subject in each example. From the left to right, the input images, results of expression removal, results of facial expression transfer.}
\label{fig:res_transfer_CK}
\end{figure}

\begin{figure}[t]
\begin{center}
\includegraphics[width=0.95\linewidth]{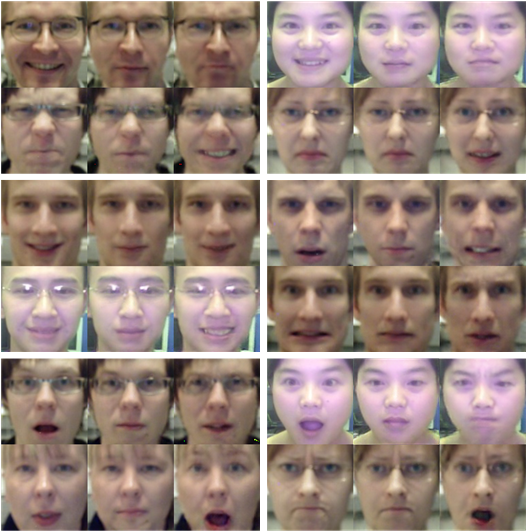}
\end{center}
   \caption{Results of Oulu-CASIA database for facial expression transfer. Images are arranged by the same order as Fig.~\ref{fig:res_transfer_CK}.}
\label{fig:res_transfer_Oulu}
\end{figure}

Fig.~\ref{fig:res_transfer_CK} and Fig.~\ref{fig:res_transfer_Oulu} show some example results.
The facial expressions are transferred between two subjects in an identity consistent way. Besides, identity-irrelevant face attributes, e.g., eyeglasses and hairs, are perfectly preserved.
Individual differences are considered in facial expression transfer, resulting in various local deformations for different subjects. For example, when different people keep the same expression of smile, more obvious changes can be discovered for people with larger mouths.

%

\subsubsection{Facial Expression Interpolation}
Interpolation for unseen expression is conducted in this experiment to demonstrate our model's capability to synthesize expressions with different intensities.
It is worth noting that there is no ground-truth in this experiment, and the locations of the fiducial points are obtained from a pre-trained shape dictionary as described in Sec.~\ref{sec:Geometry_Manipulation}.

\begin{figure}[t]
\begin{center}
\includegraphics[width=1.0\linewidth]{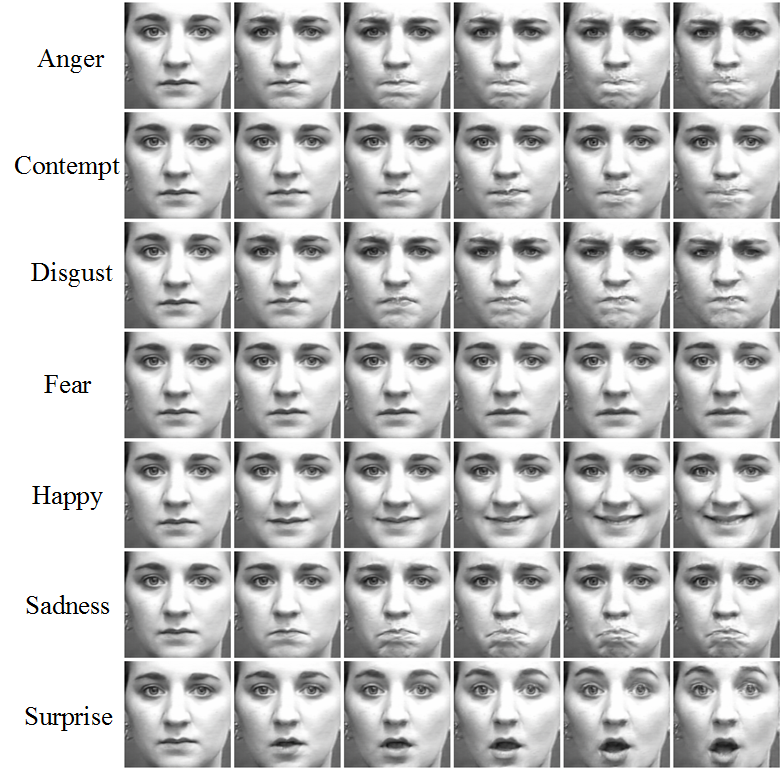}
\end{center}
   \caption{Results of CK+ database for facial expression interpolation. Images in the left-most column are the source images, and the remainder are synthesized results. Each row shows a different expression with ascending intensity from left to right. Seven expression styles are shown corresponding to the annotated expression classes in CK+ database.}
\label{fig:res_inter_CK}
\end{figure}

\begin{figure}[t]
\begin{center}
\includegraphics[width=1.0\linewidth]{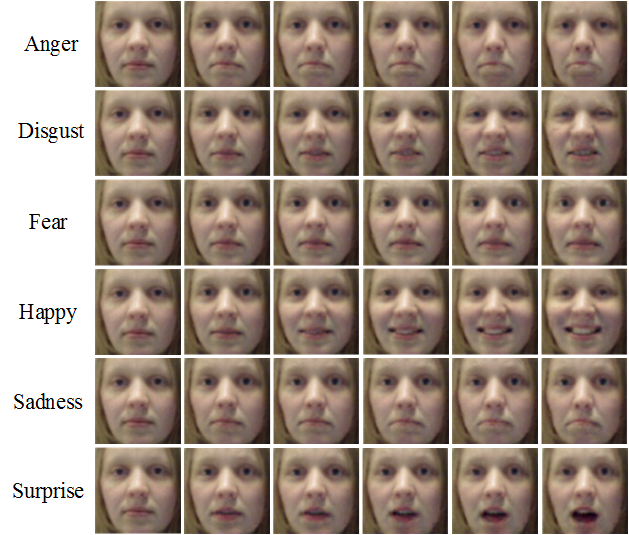}
\end{center}
   \caption{Results of Oulu-CASIA database for facial expression interpolation. Images are arranged by the same order as Fig.~\ref{fig:res_inter_CK}. Six expression styles are corresponding to the annotated expression classes in Oulu-CASIA database.}
\label{fig:res_inter_Oulu}
\end{figure}

The generated images are shown in Fig.~\ref{fig:res_inter_CK} and Fig.~\ref{fig:res_inter_Oulu}, in which each row contains a new type of expressions with different intensities. G2-GAN successfully transforms the input faces to new unseen expressions with fine details.
Especially for the results on the CK+ database, the changes of facial textures caused by expression change are well captured such as glabellar winkles under expressions of anger and disgust, chin wrinkles when mouth shut and brows lifting when scared.
This validates that the proposed G2-GAN's adjustability in generating multiple face expressions, not limited in pre-determined categories. Besides, these results also demonstrate the operation-friendliness of our method, as we can easily synthesize expressions of desired intensities.
An interesting phenomenon is that our model can distinguish the deformations of the mouth caused by happiness and surprise, and the teeth are only generated when synthesizing a smile expression.

\begin{table*}[htbp]
  \centering
  \caption{Results for expression-invariant face recognition on CK+ and Oulu-CASIA databases. Images in the probe set are processed by our expression removal model firstly, and then fed to face recognition models. We conduct face verification on the transformed probe set and the original gallery set. Results of the `original' configuration are obtained by directly testing on the non-transformed gallery set as well as probe set.}
    \begin{tabular}{|p{1.9cm}<{\centering} | p{3.2cm}<{\centering} | p{1.1cm}<{\centering} | p{1.4cm}<{\centering} |  p{1.4cm}<{\centering} | p{1.1cm}<{\centering}|  p{1.4cm}<{\centering} | p{1.4cm}<{\centering}|}
    \hline
     \multirow{2}[2]{*}{Dataset}&\multirow{2}[2]{*}{Configuration} & \multicolumn{3}{c|}{VGG-Face} & \multicolumn{3}{c|}{Light CNN} \\
     \cline{3-8}
           & & Rank-1 & FAR=1\%  & FAR=0.1\%  & Rank-1 & FAR=1\%  & FAR=0.1\% \\
    \hline
    \multirow{5}[2]{*}{CK+} & original & 96.41  & 92.13 & 88.11  & \textbf{100.00} & 97.01  & 93.33 \\
     \cline{2-8}
          & w/o $L_{cyc}$, $L_{identity}$  &  96.15 & 93.33 & 84.94  & 98.63 & 96.83  & 87.77  \\
          & w/o $L_{cyc}$  & 96.15  & 94.27 & 87.25  & \textbf{100.00} &  97.60 &  94.87 \\
          & w/o $L_{identity}$  & 96.41  & 92.90 & 84.86  & 99.23 & 97.43  & 89.65  \\
          & G2-GAN  &  \textbf{97.26} & \textbf{96.15} & \textbf{92.22}  & \textbf{100.00} & \textbf{97.69}  & \textbf{94.95} \\
    \hline
    \multirow{5}[2]{*}{Oulu-CASIA} & original  & 97.68  & 94.63 & 90.91 &  \textbf{99.92} & 95.35  & 89.02  \\
    \cline{2-8}
           & w/o $L_{cyc}$, $L_{identity}$  & 96.95  & 94.99  & 90.46  & 99.52 & 95.95  & 90.10   \\
          & w/o $L_{cyc}$  &  97.56 & 95.80 & 92.90 &  \textbf{99.92}  &  97.60 &  91.67  \\
          & w/o $L_{identity}$  & 96.59  & 95.35  & 90.02  & 99.84 &  97.04 & 89.78\\
          & G2-GAN& \textbf{97.84}  & \textbf{96.19}  &  \textbf{93.19}  &  99.88 & \textbf{97.80}  & \textbf{93.31}  \\
    \hline
    \end{tabular}%
  \label{tab:res_faceRec}%
\end{table*}%

\subsubsection{Expression-Invariant Face Recognition}
In this subsection, we apply G2-GAN in expression-invariant face recognition. The expression removal model is employed as a normalization module in face recognition, which transforms faces into neutral expression. Face verification is taken in both the CK+ dataset and the Oulu-CASIA dataset. The gallery set is selected from the first frame of each video sequences, with only one image for each subject. The probe set is made up of all the rest images in testing set.
Two released face recognition models are tested, including the VGG-FACE~\cite{parkhi2015vggface} and the Light CNN~\cite{wu2015lightened}.
The Rank-1 identification rate, true accept rates at $1\%$ and $0.1\%$ (TAR@FAR=$1\%$, TAR@FAR=$0.1\%$) are taken as evaluation metrics. In order to validate the effectiveness of $L_{cyc}$ and $L_{identity}$, we report the results of removal each one of them respectively.

Results for the expression-invariant face recognition experiment are presented in Tab.~\ref{tab:res_faceRec}.
Benefiting from the powerful representation ability of deep learning methods, VGG-FACE and Light CNN obtain high performances on the original images. However, results can be further improved by introducing our expression removal module, especially for a lower FAR.
Both the cycle consistency loss and the identity preserving loss facilitate to improve the recognition performance according to results of w/o $L_{cyc}$, w/o $L_{identity}$, and the basic setting w/o $L_{cyc},L_{identity}$.
Besides, slight drops occur when we do not use $L_{identity}$ comparing with the results of original images, suggesting the necessity of $L_{identity}$ in face editing when the face identity is expected to be preserved.

\vspace{-0.005\linewidth}
\section{Conclusions}

This paper has developed a geometry-guided adversarial framework for facial expression synthesis. Facial geometry has been employed to guide photo-realistic face synthesis as well as to provide an operation friendly solution for specifying target expression. Besides, a pair of facial editing subnetworks are trained together towards two opposite tasks: expression removal and expression synthesis, forming a mapping cycle between expressionless and expressioned faces. By combining these two subnetworks, our method can be used in many face related applications including facial expression transfer and expression-invariant face recognition. Moreover, we have proposed an individual-specific shape model for operating the facial geometry, in which individual differences are considered. Extensive experimental results demonstrate the effectiveness of the proposed method for facial expression synthesis.

%

{\small
\bibliographystyle{ieee}
\bibliography{FaceEditing_review2}
}
\begin{appendix}
\section{Supplementary Material}
In this supplementary material, we present fully detailed information on 1) comparison experiment with ExprGAN~\cite{Ding2017ExprGAN} on the Oulu-CASIA dataset; 2) expression synthesis results for fine-grained control of eye status.

\subsection{Comparison experiment with ExprGAN~\cite{Ding2017ExprGAN}}

\begin{figure*}[t]
\begin{center}
\includegraphics[width=0.97\linewidth]{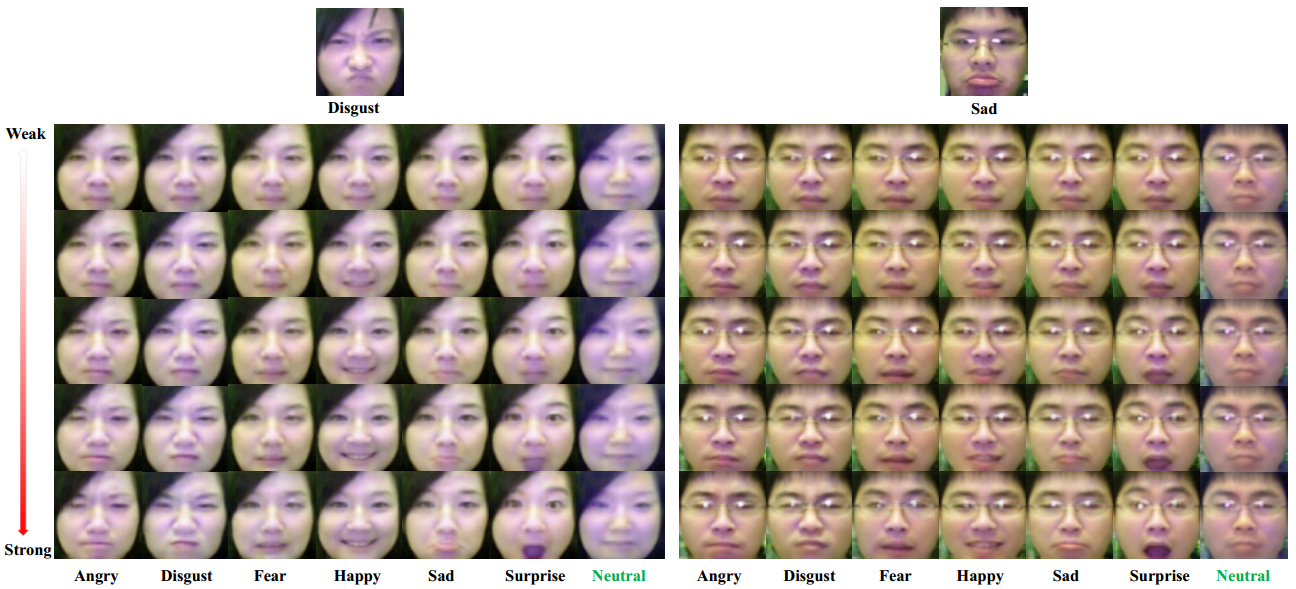}
\end{center}
   \caption{Facial expression synthesis results of ExprGAN~\cite{Ding2017ExprGAN} on the Oulu-CASIA dataset(originally shown in~\cite{Ding2017ExprGAN}).}
\label{fig:res_exprgan1}
\end{figure*}

\begin{figure*}[t]
\begin{center}
\centering
\subfigure[P076]{\includegraphics[width=0.48\textwidth]{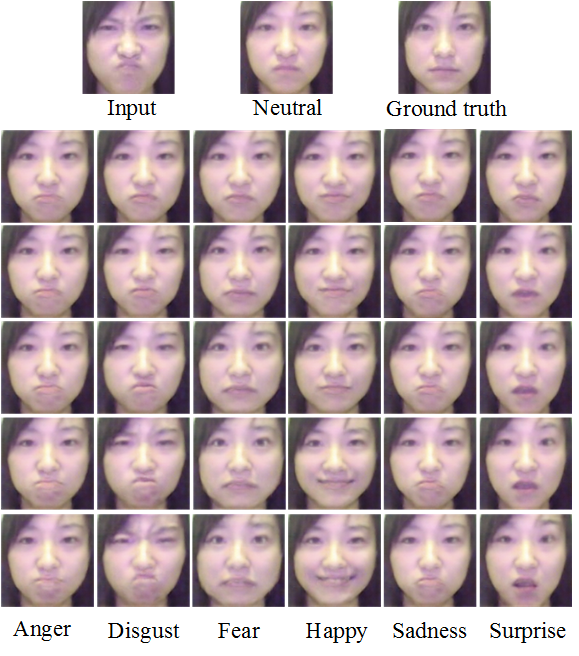}\label{fig:res_syn_ours_076}}
\subfigure[P074]{\includegraphics[width=0.48\textwidth]{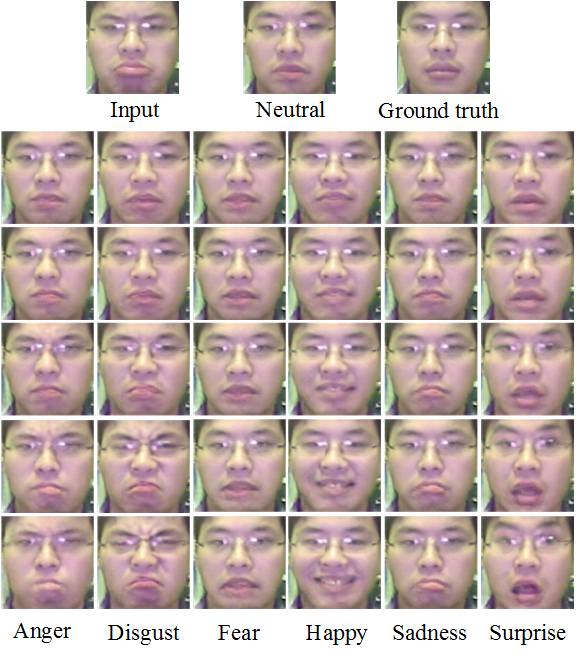}\label{fig:res_syn_ours_074}}
\end{center}
   \caption{Facial expression synthesis results of the proposed G2-GAN. We take the same source images with ExprGAN~\cite{Ding2017ExprGAN} for comparison. The `neutral' face image is generated from the `input' face. }
\label{fig:res_syn_ours}
\vspace{-2mm}
\end{figure*}

As mentioned in our paper, ExprGAN is the most similar work to ours. In this part,
we take the same facial expression synthesis experiments with ExprGAN~\cite{Ding2017ExprGAN} for comparison.

Fig.~\ref{fig:res_exprgan1} shows the results of ExprGAN on the Oulu-CASIA dataset~\cite{chen2009learning}. Images in the top row are input faces, and the rest are synthesized expressions with ascending intensities from top to down. Six types of expressions are synthesized, which correspond to the annotated expression classes in Oulu-CASIA dataset respectively. In addition, the neutral expression faces can also be generated in ExprGAN.
In order to compare with ExprGAN, we take the same expression synthesis experiments and use the same input images with ExprGAN as shown in Fig.~\ref{fig:res_syn_ours}. Firstly, we employ the proposed expression removal model to recover the neutral expression faces, which are shown in the top row. Then, expression synthesis is conducted on neutral expression faces to generate various expressioned faces. Following the setting in ExprGAN, we synthesize six types of expressions with five different intensities. It is worth noting that these two input images (subject id in the Oulu-CASIA dataset are P074 and P076 respectively) are not in our training dataset.

Due to the different way of image cropping, larger face areas (especially the chin areas) are covered in our experiments than ExprGAN. Particularly, the chin areas show wide variations along with expression changes, resulting in more difficulties in learning expression synthesis model. Comparing with ExprGAN, G2-GAN does much better in preserving identity information and keeping local details (such as hair in Fig.~\ref{fig:res_syn_ours_076} and beard in Fig.~\ref{fig:res_syn_ours_074}) through the expression transformation process. The neutral faces recovered by ExprGAN do not look like the ground truth images that are shown in the top row of Fig.~\ref{fig:res_syn_ours}, wheras G2-GAN is able to generate neutral faces without losing much identity information. Besides, we can see that the proposed G2-GAN can generate expressioned faces with fine details such as wrinkles caused by frown and pout, whereas the results generated by ExprGAN tend to lack these details.

\subsection{Expression synthesis with controlled eye status}

The usage of face geometry in our framework provides an intuitive way for specifying target facial expression, with which we can generate special expressions such as ``a lopsided grin with one eye open". In this part, we show the ability of G2-GAN to synthesize facial expressions with controlled status of eye (open and closed).
Since images in the Oulu-CASIA dataset are of low-resolution, we only take this experiment on the CK+ dataset~\cite{lucey2010extended}.

\begin{figure*}[t]
\begin{center}
\includegraphics[width=0.85\linewidth]{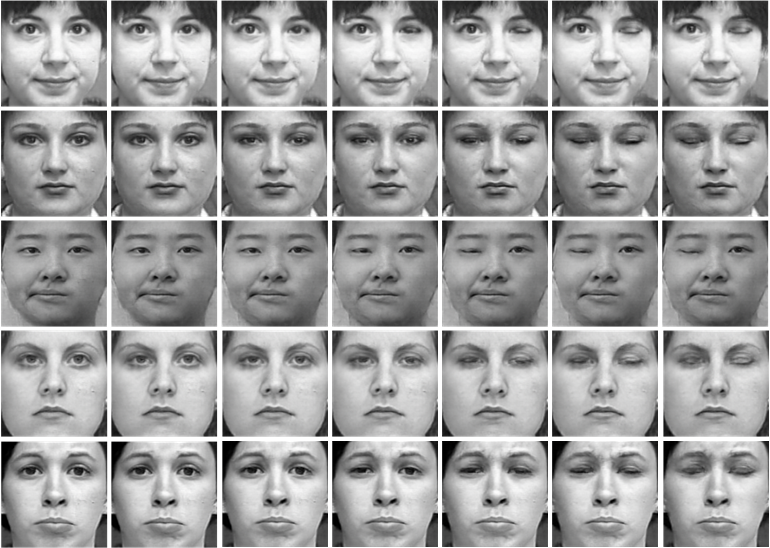}
\end{center}
   \caption{Examples of synthesized facial expressions with controlled status of eye. From left to right, eyes are gradually closed.}
\label{fig:res_close_eye}
\end{figure*}

\begin{figure*}[t]
\begin{center}
\includegraphics[width=0.85\linewidth]{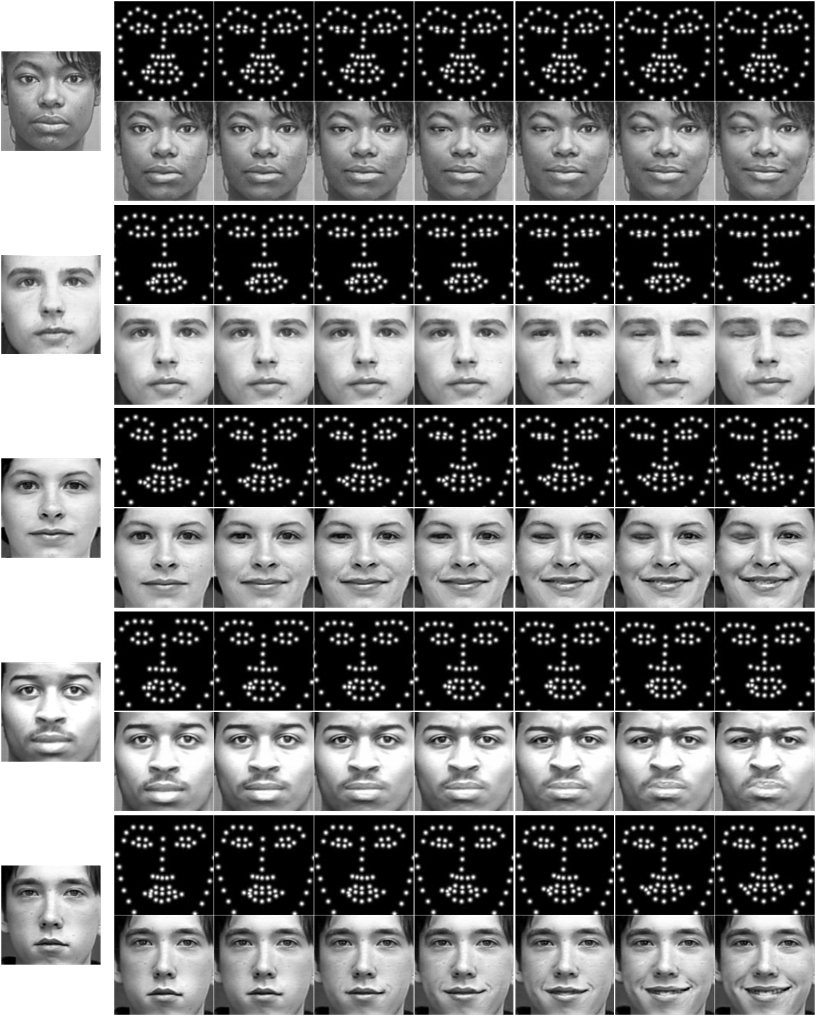}
\end{center}
   \caption{Examples of synthesized facial expressions on the CK+ dataset. Images in the first column are input faces, and the rest are input heatmaps and synthesized results.}
\label{fig:res_heat}
\end{figure*}

\begin{figure*}[t]
\begin{center}
\includegraphics[width=0.85\linewidth]{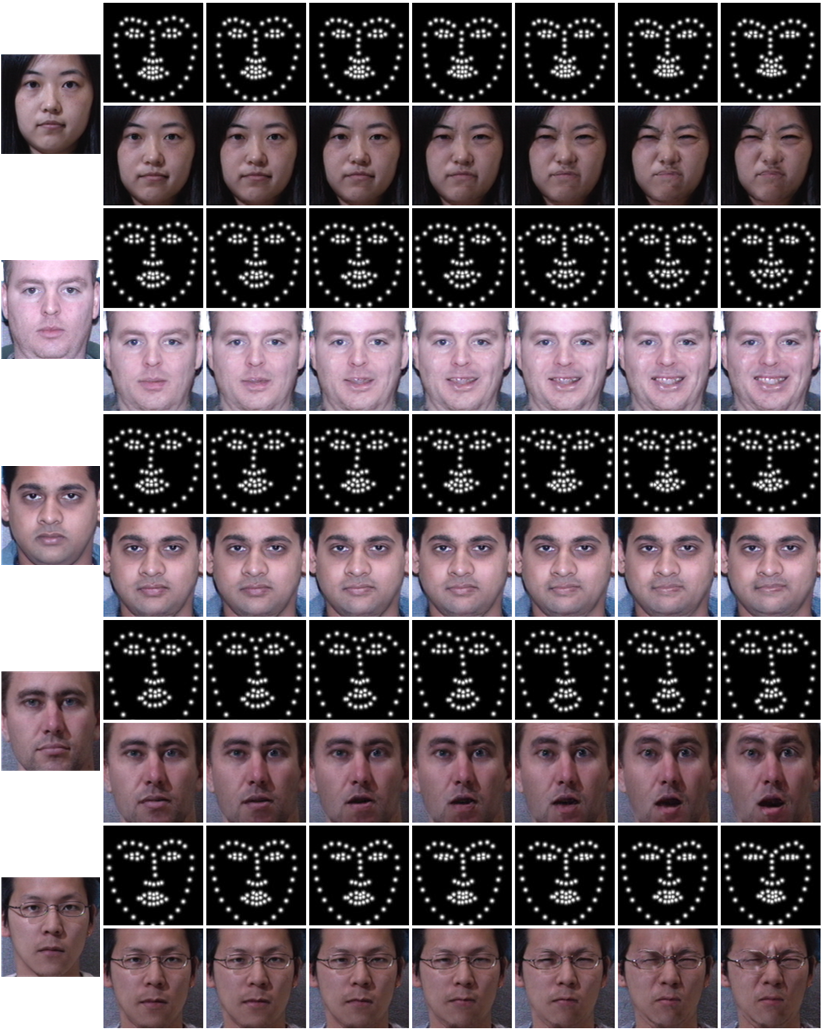}
\end{center}
   \caption{Examples of synthesized facial expressions on the MultiPIE dataset. Images in the first column are input faces, and the rest are input heatmaps and synthesized results.}
\label{fig:res_heat_mpie}
\end{figure*}

Fig.~\ref{fig:res_close_eye} shows examples of synthesized images with various eye statues. We can see that the proposed G2-GAN not only generates compelling perceptual results but also preserves identity information well. By directly manipulate the face geometry, we can perform fine-grained control of the eye status, which is hard for other generative model-based approaches.
These results demonstrate the superiority of G2-GAN in operation-friendliness as well as the diversity of synthesized faces, suggesting potential applications for face edit. More results on the CK+ dataset and MultiPIE dataset~\cite{Gross2008Multi} can be found in Fig.~\ref{fig:res_heat} and Fig.~\ref{fig:res_heat_mpie}.

\end{appendix}

\end{document}